\documentclass[runningheads]{llncs}
\usepackage{graphicx}
\usepackage{tikz}
\usepackage{comment}
\usepackage{amsmath,amssymb} 
\usepackage{color}

\usepackage[T1]{fontenc} 

\usepackage[pagebackref,breaklinks,colorlinks]{hyperref}

\begin{document}

\pagestyle{headings}
\mainmatter
\def\MMMSubNumber{232}  

\title{Marine Video Kit: A New Marine Video Dataset for Content-based Analysis and Retrieval}

\titlerunning{Marine Video Kit}
%

\author{Quang-Trung, Truong\inst{1} \and	
Tuan-Anh Vu\inst{1} \and	
Tan-Sang Ha\inst{1} \and	
Jakub Loko\v{c}\inst{2} \and	
Yue Him Wong Tim\inst{3} \and	
Ajay Joneja\inst{1} \and	
Sai-Kit Yeung\inst{1}	
}

\authorrunning{Truong et al.}
\institute{Hong Kong University of Science and Technology \and
FMP, Charles University, Prague, Czech Republic \and
Shenzhen University
}

\maketitle              
\begin{abstract}
Effective analysis of unusual domain specific video collections represents an important practical problem, where state-of-the-art general purpose models still face limitations. Hence, it is desirable to design benchmark datasets that challenge novel powerful models for specific domains with additional constraints. It is important to remember that domain specific data may be noisier (e.g., endoscopic or underwater videos) and often require more experienced users for effective search. In this paper, we focus on single-shot videos taken from moving cameras in underwater environments, which constitute a nontrivial challenge for research purposes. The first shard of a new Marine Video Kit dataset is presented to serve for video retrieval and other computer vision challenges. Our dataset is used in a special session during Video Browser Showdown 2023. In addition to basic meta-data statistics, we present several insights based on low-level features as well as semantic annotations of selected keyframes. The analysis also contains experiments showing limitations of respected general purpose models for retrieval. Our dataset
and code are publicly available at \url{https://hkust-vgd.github.io/marinevideokit}.
 
\end{abstract}
%

\begin{figure*}
\begin{center}
    \centering
\includegraphics[width=1\textwidth]{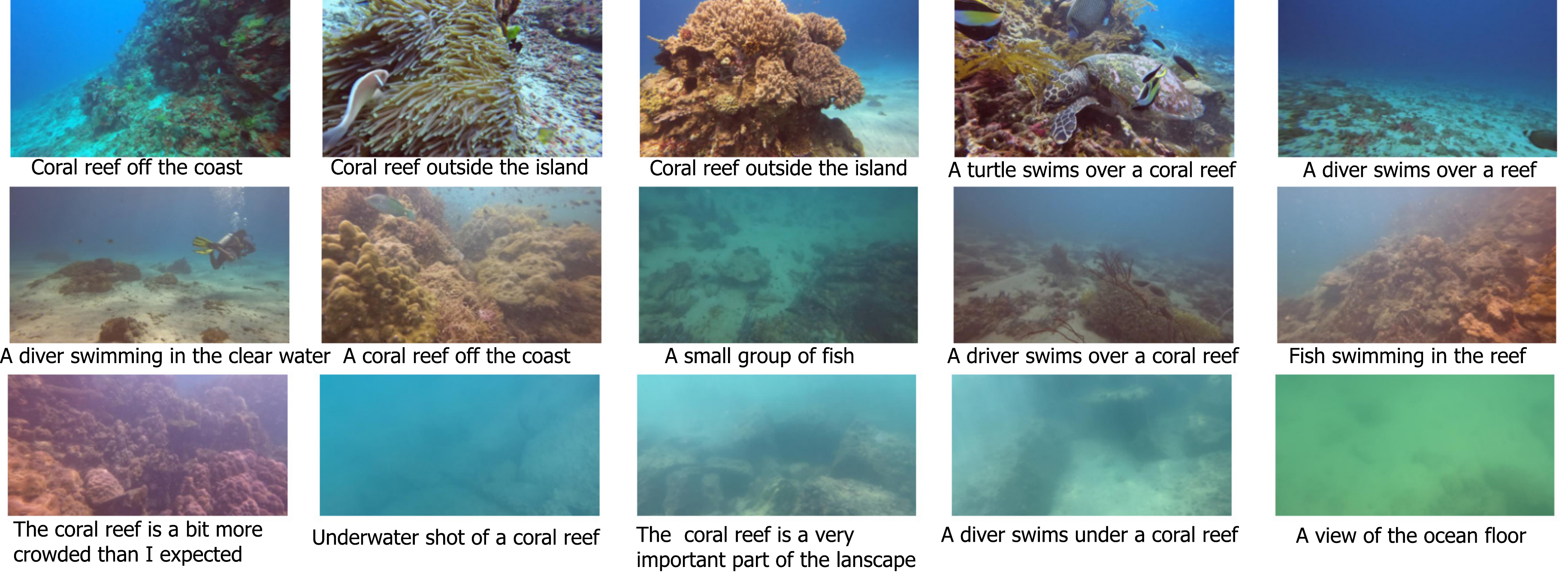}
\end{center}
  \caption{Several examples of dataset video frames and their ClipCap descriptions.}
\label{fig:captioningExamples}
\end{figure*}

\section{Introduction}
\label{sec:introduction}
In order to facilitate the development of multimedia retrieval and analysis models, the research community establishes and uses various benchmark multimedia datasets \cite{krishna2017dense,xu2016msr,ZhouCVPR18,mithun2018learning,Youngjae2018eccv,chen2018temporally,krishna2017dense,xu2016msr,ZhouCVPR18}. The datasets usually provide so-called ground-truth annotations and allow repeatable experimental comparison with state-of-the-art methods.

The source of large benchmark collections is often a video sharing platform (e.g., Youtube, or Vimeo) with specific licensing of the content. For example, the respected Vimeo Creative Commons collection dataset (V3C) \cite{rossetto2019v3c} contains several thousand hours of videos downloaded from the Vimeo platform. Although only videos with the creative commons license were used for the dataset, it is available only with an agreement form indicating possible changes in the future. Hence, it is beneficial to design also new video datasets with a limited number of copyright owners, limiting potential future changes to the dataset (experiment repeatability). Designing datasets with highly challenging content is also necessary, even for a limited number of data items. This aspect is especially important for interactive search evaluation campaigns \cite{GurrinZH0DLTHRS22,HellerGBG0LLMPR22} addressing a broad community of researchers from different multimedia retrieval areas. Indeed, for many research teams (especially smaller ones), it might be more feasible to participate in a difficult challenge over 10-100 hours of videos, rather than a challenge over 10.000 hours and more.

In general everyday videos, there appear many common classes of objects, and thus even a larger collection can be effectively filtered with text queries. On the other hand, a domain specific collection (i.e., one cluster with a lower variance of common keywords) might already be challenging for lower sizes of collections. Therefore, we selected underwater marine videos with the seafloor, coral reefs, and various biodiversity where potentially effective keywords are unknown to ordinary users. Furthermore, unlike common ``everyday'' videos, underwater videos pose additional obstacles for multimedia analysis and retrieval models.
 These challenges include low visibility, blurry shots, varying sizes and poses of objects, a crowded background, light attenuation, and scattering, among others \cite{Levy2018}.
Therefore, not only general purpose models but also domain specific classifiers require novel ideas and breakthroughs to reach human-level accuracy.

This paper presents a first fragment of a new ``Marine Video Kit'' dataset, currently intended mainly for content-based retrieval challenges. It is composed of more than 1300 underwater videos from 36 locations worldwide and at different times across the year. Whereas the long-term ambition for this dataset might be even a video-sharing platform with controllable extensions, for now, we present a manually organized set of directories comprising videos, selected frames, and various forms of meta-data. So far, the meta-data comprises available video attributes (location, time) and pre-computed captions as well as embeddings for the set of selected video frames. In the future, we plan to provide also more annotations for object detection, semantic segmentation, object tracking, etc.

The remainder of this paper is structured as follows:  section \ref{sec:related_works} gives an overview of related works, section \ref{sec:marine_kit} introduces the details of the Marine Video Kit dataset, and finally, section \ref{sec:conclusion} concludes the paper.

\section{Related Work}
\label{sec:related_works}

Numerous datasets were created for the object detection and segmentation task in order to better comprehend marine life and ecosystems. In this section, we briefly review some recent works for Marine-related datasets.

The Brackish \cite{pedersen2019brackish} is an open-access underwater dataset containing annotated image sequences of starfish, crabs, and fish captured in brackish water with varying degrees of visibility. The videos were divided into categories according to the primary activity depicted in each one. A bounding box annotation tool was then used to manually annotate each category's 14,518 frames, producing 25,613 annotations in total. 

MOUSS dataset \cite{mouss2018dataset} is gathered by a horizontally-mounted, grayscale camera that is placed between 1 and 2 meters above the sea floor and is illuminated solely by natural light. In most cases, the camera remains stationary for 15 minutes at a time in each position. There are two sequences in the MOUSS datasets: MOUSS seq0 and MOUSS seq1. The MOUSS seq0 includes 194 images, all of which belong to the Carcharhini-formes category, and each image has a resolution of 968 by 728 pixels. There is only one category in the MOUSS seq1, which is called Perciformes. Each image has a resolution of 720 by 480 pixels. A human expert was responsible for assigning each of the species labels. 

WildFish \cite{zhuang2018wildfish} is a large-scale benchmark for fish recognition in the wild. It consists of 1,000 fish categories and 54,459 unconstrained images. This benchmark was developed for the field of the classification task. In the field of image enhancement, the database known as Underwater ImageNet \cite{fabbri2018icra} is made up of subsets of ImageNet \cite{imagenet} that contain photographs taken underwater. As a result, the distorted and undistorted sets of underwater images were allowed to have 6,143 and 1,817 pictures, respectively. Fish4Knowledge \cite{fisher2016fish4knowledge} gave an analysis of fish video data. 

OceanDark \cite{porto2019} is a novel low-lighting underwater image dataset that was created to quantitatively and qualitatively evaluate the proposed framework. This dataset was developed in the field of image enhancement and consisted of images that were captured using artificial lighting sources.

The Holistic Marine Video Dataset \cite{hmv2021}, also known as HMV, is a long video that simulates marine videos in real time and annotates the frames of HMV with scenes, organisms, and actions. The goal of this dataset is to provide a large-scale video benchmark with multiple semantic aspect annotations. On the other hand, they also provide baseline experiments for reference on HMV for three tasks: the detection of marine organisms, the recognition of marine scenes, and the recognition of marine organism actions.

\section{Marine Video Kit Dataset}
\label{sec:marine_kit}
In this section, we will show details about our Marine Video Kit Dataset, which provides a dense, balanced set of videos focusing on marine environment, hence enriching the pool of existing marine dataset collections.

Many types of cameras were used to build the presented first shard of the Marine Video Kit dataset, such as Canon PowerShot G1 X, Sony NEX-7, OLYMPUS PEN E-PL, Panasonic Lumix DMC-TS3, GoPro cameras, and consumer cellphones cameras. The dataset consists of a larger number of single-shot videos without post-processing. Unlike common video collections taken by crawling search engines, the presented dataset focuses solely on marine organisms captured during diving periods. The typical duration of each video is about 30 seconds.

To illustrate the specifics of the underwater environment, we present 3D color histograms showing differences for different parts of the dataset. 
For a larger set of selected frames (extracted with 1fps), we also analyze semantic descriptions automatically extracted by the ClipCap model \cite{mokady2021clipcap}. For a randomly selected subset of 100 frames, the automatically generated descriptions are compared with manually created annotations in a known-item search experiment.

\begin{figure*}
\begin{center}
\centering
\includegraphics[width=.99\textwidth]{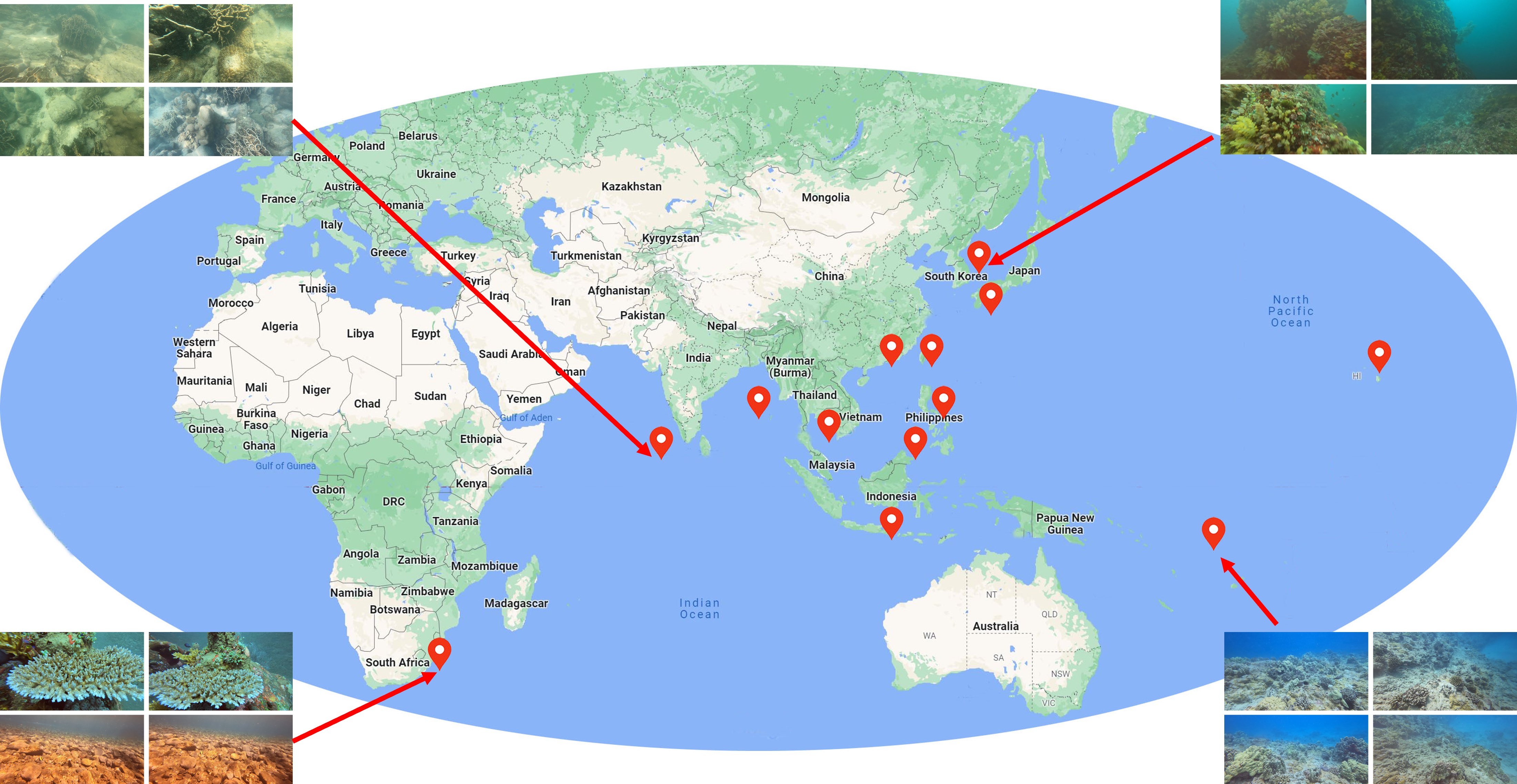}
\end{center}
  \caption{The world map illustrates countries/regions where we capture data around the world (map source: Google Maps). }
\label{fig:map}
\end{figure*}

\subsection{Acquiring the Dataset}

Different ocean areas have their own marine biodiversities, such as various animals, plants, and microorganisms. Therefore, to create a diverse dataset for the research community, we have visited and captured data from 11 different regions and countries during daytime and nighttime (unless bad weather or other circumstances happen). We present a world map illustrating 11 countries/regions where the videos were captured Fig.~\ref{fig:map}. 

We categorize the captured videos in terms of their location and time, then utilize OpenCV library \cite{opencv_library} to process all data into one unifying format (JPG for images, MP4 for videos). Due to the variety of capturing devices, our raw videos have different resolutions, from the HD (720p) resolution to Ultra HD (4K), with a frame rate of 30 fps. Therefore, we also utilize FFmpeg library\cite{tomar2006converting} to convert all data to low and high resolution for different research purposes, such as video retrieval, super-resolution, object detection, segmentation, etc.

\subsection{Dataset Structure}

\begin{figure*}
\begin{center}
    \centering
    \includegraphics[width=0.6\textwidth]{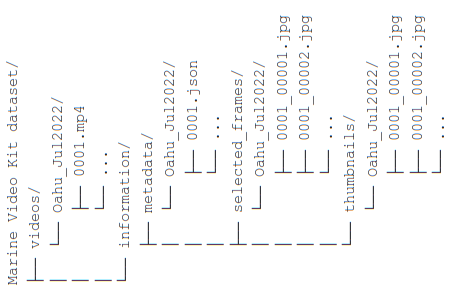}
\end{center}
    \caption{Directory structure of the Marine Video Kit dataset. }
    \label{fig:datastructure}
\end{figure*}

In this section, we show our data's directory structure, which organizes different aspects of the data. As shown in Fig. ~\ref{fig:datastructure}, there are two sub-directories for video and its supplementary information. 

For each video directory, we format their name as \textit{location\_time} pattern to explicitly represent the time and location that they were captured. For example, ``Oahu\_Jul2022'' was captured at Oahu - the third-largest of the Hawaiian Islands, in July 2022. 

For each information directory, the \textit{selected\_frames} directory stores all frames that are evenly selected one frame per second and are kept in the original resolution, while the \textit{thumbnails} directory stores the same frame but in down-scaled resolution. Finally, the \textit{metadata} directory contains the associated meta information of each video in JSON format for easier sharing and parsing information. This metadata file contains semantic and statistical information, such as video name, duration, height, width, camera device, directory, license, and reference information, such as ClipCap captions. 

\subsection{Dataset statistics}

\begin{figure*}
\begin{center}
\centering
\includegraphics[width=1\textwidth]{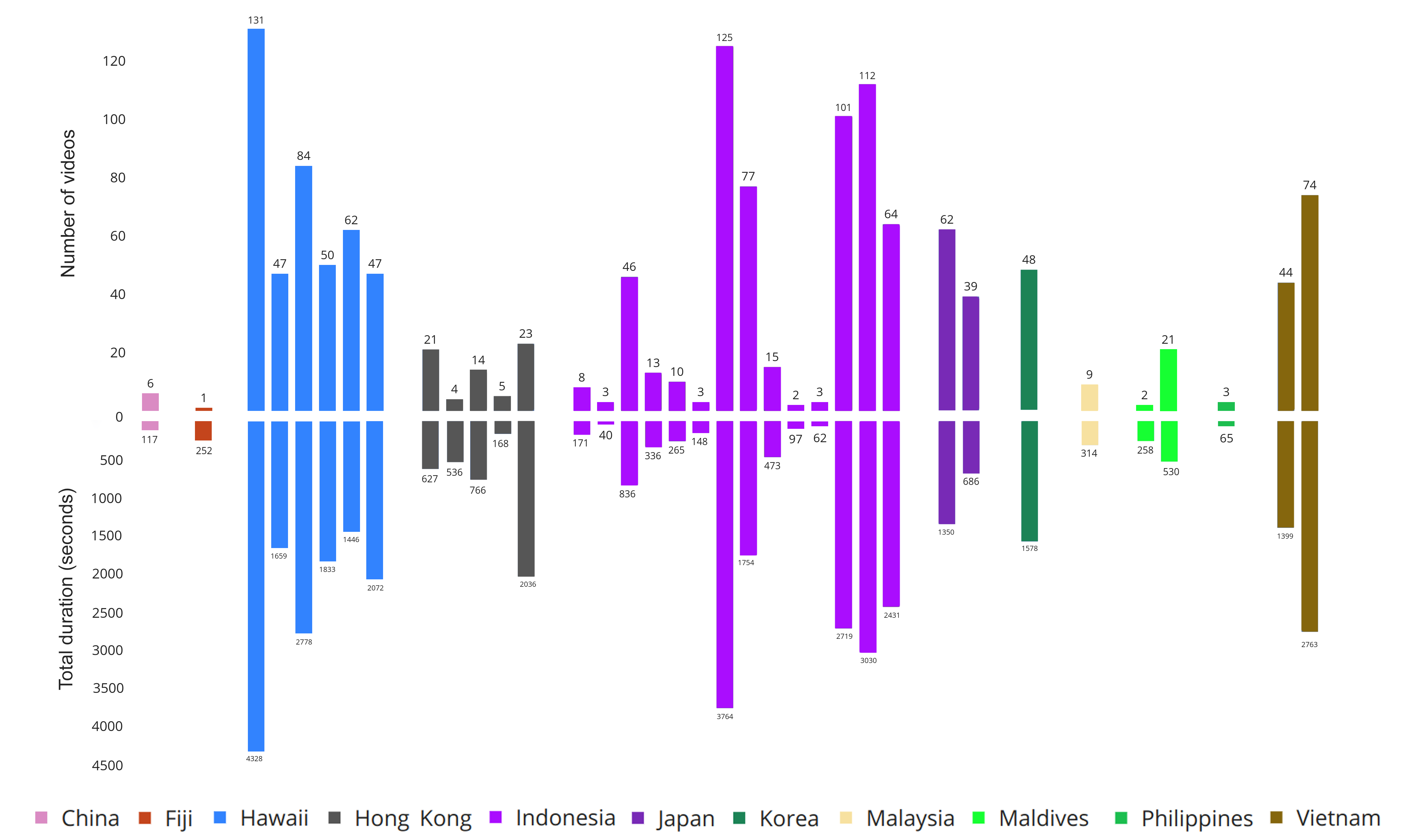}
\end{center}
  \caption{The figure shows the number of videos and overall time duration for each region.}
\label{fig:categorynum}
\end{figure*}

We capture data from 11 different regions and countries during the time from 2011 to 2022. There is a total of 1379 videos with a length from 2 seconds to 4.95 minutes, with the mean and median duration of each video is 29.9 seconds, and 25.4 seconds, respectively. The total duration is slightly above 12 hours, however, the diving time is significantly larger, up to a thousand hours. Fig. ~\ref{fig:categorynum} shows the number of videos, and the total length of videos varies by category. 

\begin{figure*}
\begin{center}
\centering
\includegraphics[width=1\textwidth]{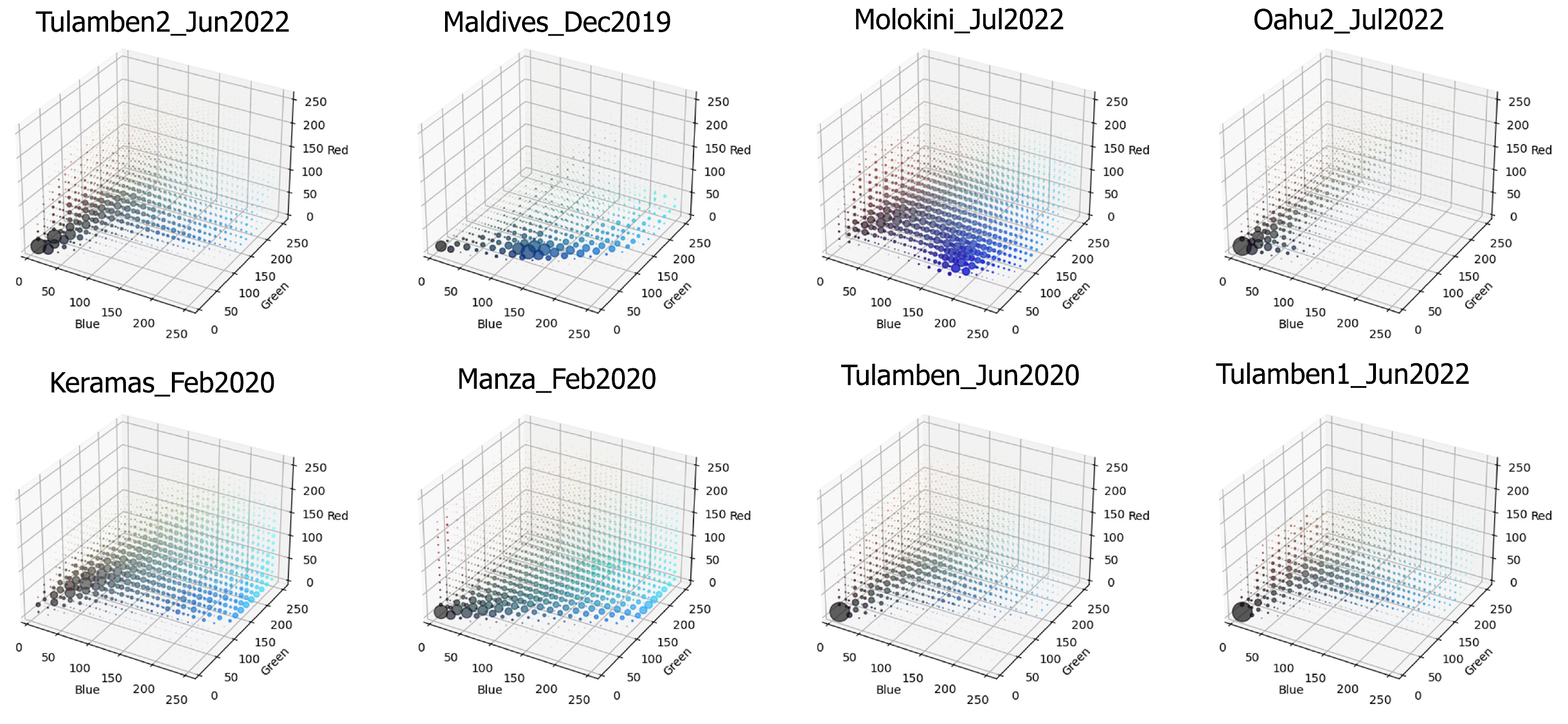}
\end{center}
  \caption{Illustration of video frame colors present in regions under low illumination using 3D color histograms.}
\label{fig:3dhistogram_dark}
\end{figure*}

\begin{figure*}
\begin{center}
\centering
\includegraphics[width=0.85\textwidth]{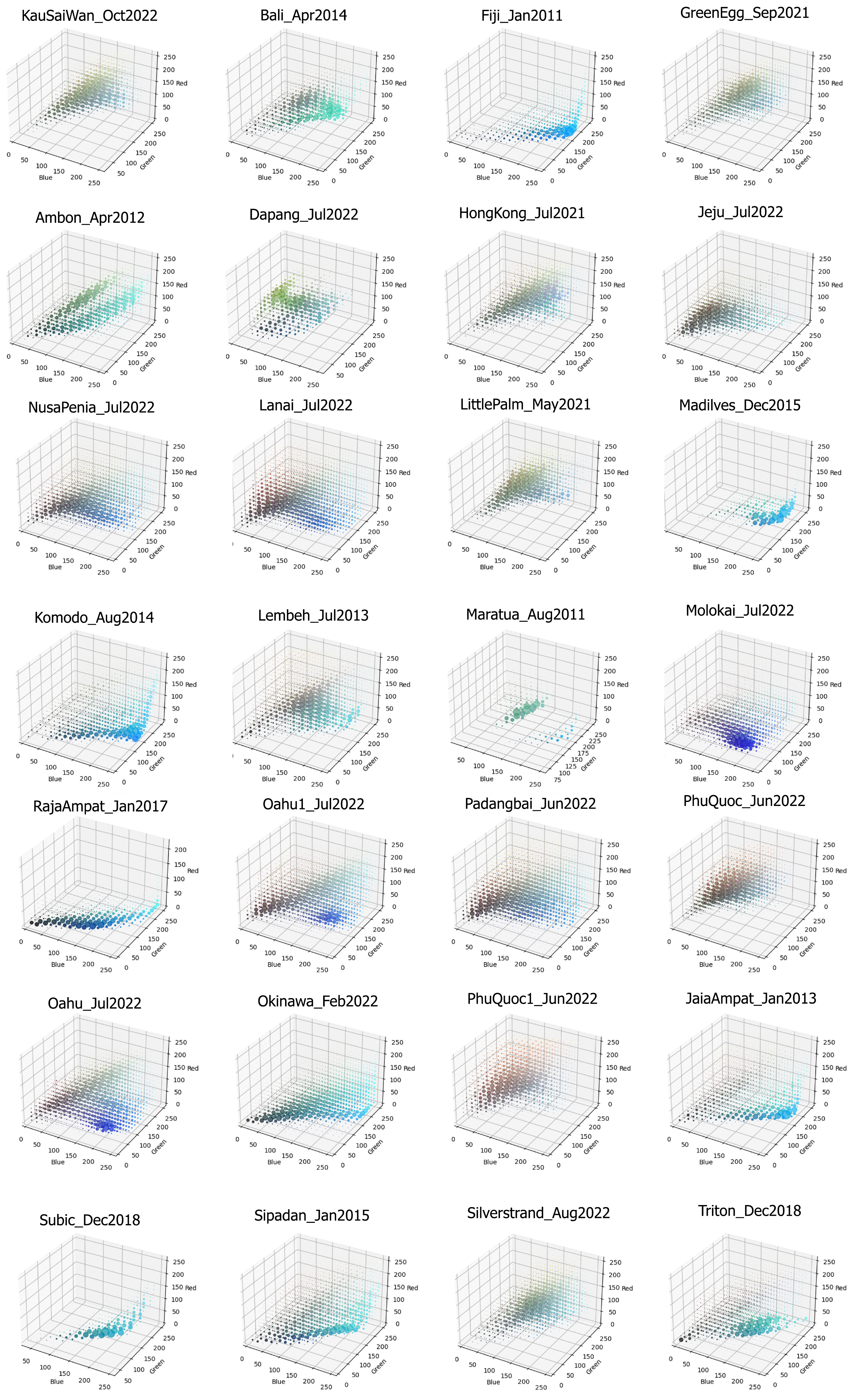}
\end{center}
  \caption{Illustration of video frame colors present in regions under good illumination using 3D color histograms.}
\label{fig:3dhistogram_lighting}
\end{figure*}

To represent marine videos from directories in color space, 3D color histograms are used to aggregate pixels to a $256\times256\times256$ array for 3D representation. The 3D visualization of each directory is presented in Fig. ~\ref{fig:3dhistogram_dark}, ~\ref{fig:3dhistogram_lighting} with 3D points colored by RGB histogram values based on corresponding places in 3D spaces, and the sizes are proportional to the color densities. We used color quantization and normalization for a more informative illustration of 3D color histogram densities.

A big challenge when using Marine Video Kit dataset for retrieval tasks is that the marine environment has changed considerably, and the captured data is heavily affected by lighting variations or caustics, thus causing low visibility. Fig. ~\ref{fig:3dhistogram_dark} illustrates regions under low lighting, and Fig. ~\ref{fig:3dhistogram_lighting} mentions regions under good lighting.

\subsection{ClipCap Captions}

\begin{figure*}
\begin{center}
\centering
\includegraphics[width=.8\textwidth]{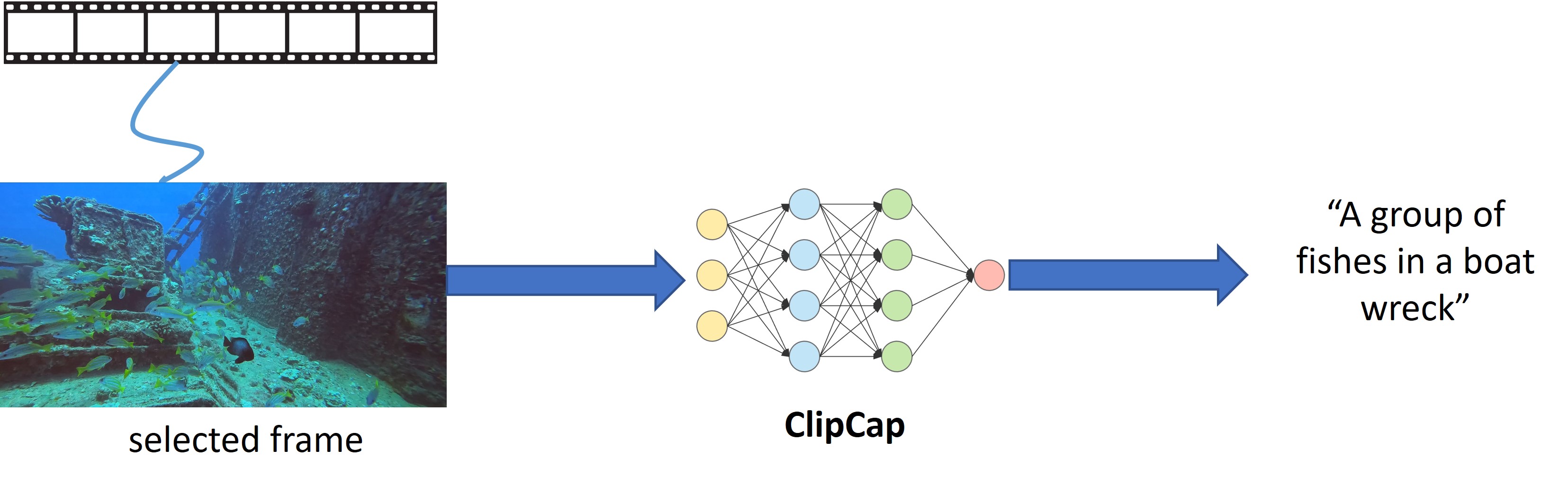}
\end{center}
  \caption{The captioning architecture. A selected frame is extracted from marine video, then fed into the ClipCap model to output the caption of the selected frame.}
\label{fig:captioning}
\end{figure*}

To provide semantic information for the set of uniformly selected frames, ClipCap \cite{mokady2021clipcap} architecture was employed. ClipCap elegantly combines rich features of the CLIP model \cite{radford2021learning} with the powerful GPT-2 language model \cite{li2021prefix}. Specifically, a prefix is computed for the image feature vector, while the language model continues to generate the caption text based on the prefix. Additionally, ClipCap is efficient in diverse datasets, which motivates us to caption selected frames in the marine dataset. Captions given by ClipCap are utilized as an automatically generated caption attribute in video metadata files.

\begin{figure*}
\begin{center}
    \centering
\includegraphics[width=1\textwidth]{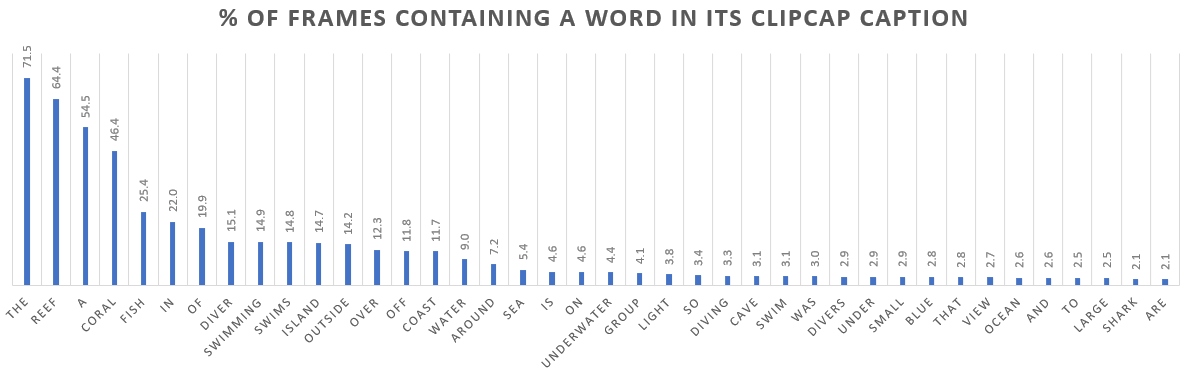}
\end{center}
  \caption{Occurrence of words in frame captions. Computed for a subset of the dataset.}
\label{fig:Words}
\end{figure*}

We adopt ClipCap for selected frames in Fig. ~\ref{fig:captioning} that are mentioned in metadata files. The process to generate captions consists of two steps to ensure content relevance. The captions of selected frames are automatically outputted by the ClipCap model; after that, we regularly inspect the captions to remove unrelated descriptions. Fig.  ~\ref{fig:captioningExamples} shows selected frame captions of ClipCap that are utilized to describe semantic information of the marine dataset, and Fig. ~\ref{fig:Words} presents frequencies for individual words in frame captions.

\subsection{Known-item Search Experiment}
In this section, we show that the new dataset represents a challenge for an information retrieval task. Specifically, a known-item search experiment is performed using a subset of all selected video frames $O_i$ (extracted with 1fps, about 40K frames), and their CLIP representations \cite{CLIP}. The experiment assumes a set of pairs $[q_i, t_i]$, where $t_i$ is a CLIP embedding vector representing a target (searched) video frame $O_i$, and $q_i$ is a CLIP embedding vector representing a query description of the target image $O_i$.
Using one pair, it is possible to rank all selected video frames based on their cosine distance to the query vector $q_i$. From this ranking, the rank of the target image $O_i$ can be identified and stored. Repeating this experiment for all available pairs $[q_i, t_i]$, a distribution of ranks of searched items can be analyzed.

\begin{figure*}
\begin{center}
\centering
\includegraphics[width=0.8\textwidth]{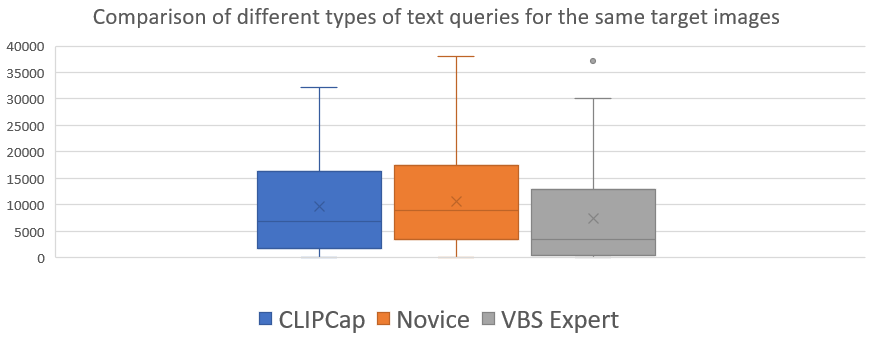}
\end{center}
  \caption{Ranks for ClipCap, Novice, and VBS Expert text queries for 100 target images. }
\label{fig:ClipCapManual}
\end{figure*}

Figure  \ref{fig:ClipCapManual} shows the result of our preliminary KIS experiment, where 100 pairs $[q_i^{novice}, t_i]$ and 100 pairs $[q_i^{expert}, t_i]$ were created manually. Novice and VBS Expert annotators described randomly selected target images, where the VBS Expert is not an expert in the marine domain but has experience with query formulation from the Video Browser Showdown. ClipCap annotations were computed for the same target images as well. Using CLIP embeddings, ranks of target images were computed for the novice, VBS Expert, and ClipCap queries. Figure \ref{fig:ClipCapManual} shows that the overall distributions of ranks are similar for ClipCap and novice users, while the VBS expert was able to reach a better average and median rank. In the future, we plan more thorough experiments.
We also note that employed random selection of target images often does not lead to unique items, and users did not see the search results for the provided 100 text descriptions/queries. Both notes affect the overall distribution of ranks.

\begin{figure*}
\begin{center}
    \centering
\includegraphics[width=1\textwidth]{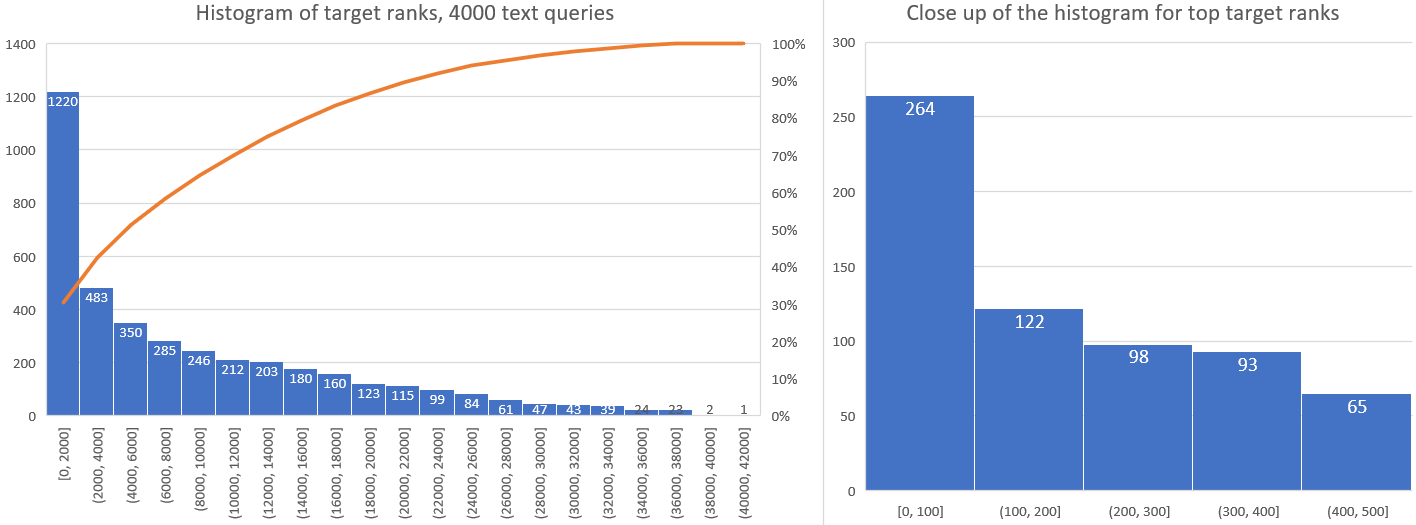}
\end{center}
  \caption{4000 KIS queries using ClipCap captions, histogram bins aggregate ranks of target images for the corresponding queries.}
\label{fig:KIS4000}
\end{figure*}

Since ClipCap descriptions are available for all database frames, we also present another KIS experiment for ClipCap based queries with similar average target rank as novice users in Figure \ref{fig:ClipCapManual}.
Figure \ref{fig:KIS4000} was evaluated for 4000 pairs $[q_i, t_i]$, where the target image descriptions were obtained using the ClipCap approach. We may observe that only about 30\% of all queries are allowed to find the target in the top-ranked 2000 dataset items. Even for the small dataset. Although the CLIP retrieval model helps, it is indeed a way more difficult challenge than searching common videos with a large number of different concepts (e.g., compared to the results of a similar study here \cite{LokocS21}). Furthermore, searching top 2000 items is not trivial. Looking at the top 500 close-ups of the histogram, only 6.6\% of target items can be found in the top 100. Hence, we conclude that limited vocabulary (i.e., ClipCap like queries) and similar content make the known-item search challenge difficult for the proposed dataset even with the respected CLIP model.

\section{Conclusion}
\label{sec:conclusion}
We provide the Marine Video Kit dataset, single-shot videos challenging for content-based analysis and retrieval. To provide a first insight of the new marine dataset, basic statistics based on meta-data, low-level color descriptors, and ClipCap semantic annotations are presented. We present a baseline retrieval study for the Marine Video Kit dataset to emphasize the domain's specificity. Our experiments show that the similarity of content in the dataset causes difficulties for a respected cross-modal based know-item search approach. We hope that our dataset with the baseline for content-based retrieval will accelerate considerable progress in the marine video retrieval area.  

We plan to extend the dataset in the future with new annotations and videos from new environments. Besides, additional computer vision tasks over the dataset, such as semantic segmentation or object detection, could be prepared for the research community. Furthermore, motion analysis, fish counting, or detection tasks are also meaningful information for retrieval applications.

\paragraph{\textbf{Acknowledgements.}} 
This research project is partially supported by an internal grant from HKUST (R9429), the Innovation and Technology Support Programme of the Innovation and Technology Fund (Ref: ITS/200/20FP), the Marine Conservation Enhancement Fund (MCEF20107), Charles University grant (SVV-260588), and the Innovation Team Project of Universities in Guangdong Province (No. 2020KCXTD023).

\paragraph{\textbf{Disclaimer.}} 
Any opinions, findings, conclusions, or recommendations expressed in this material do not necessarily reflect the views of HKLTL, CAPCO, HK Electric, and the Marine Conservation Enhancement Fund.

\bibliographystyle{splncs04}
\bibliography{egbib.bib}
\end{document}